\documentclass[journal]{IEEEtran}
\ifCLASSINFOpdf
\usepackage[pdftex]{graphicx}
\else
\fi

\usepackage{listings}

\usepackage{cancel}

\usepackage{color}

\newcommand{\DigitOneBlue}[1]{\textcolor[RGB]{76,201,255}{#1}}

\newcommand{\DigitOneRed}[1]{\textcolor[RGB]{248,126,74}{#1}}

\begin{document}
	%
	\title{Cause and Effect: Hierarchical Concept-based Explanation of Neural Networks}
	%
	%
	%
	
	\author{Mohammad Nokhbeh Zaeem$^{1}$ and Majid Komeili$^{2}$
		\thanks{$^{1}$ {\tt\small mohammadnokhbehzaeem@carleton.ca}}
		\thanks{$^{2}$Majid Komeili is with School of Computer Science, Carleton University, 1125 Colonel By Drive, Ottawa, Canada.        {\tt\small majid.komeili@carleton.ca}}%
	}
	
	%
	%

	\markboth{Preprint}%
	{Preprint}
	%



	\maketitle
	
	\begin{abstract}
		In many scenarios, human decisions are explained based on some high-level concepts.
		In this work, we take a step in the interpretability of neural networks by examining their internal representation or neuron's activations against concepts.
		A concept is characterized by a set of samples that have specific features in common.
		We propose a framework to check the existence of a causal relationship between a concept (or its negation) and task classes.
		While the previous methods focus on the importance of a concept to a task class, we go further and introduce four measures to quantitativTITLEely determine the order of causality.
		Moreover, we propose a method for constructing a hierarchy of concepts in the form of a concept-based decision tree which can shed light on how various concepts interact inside a neural network towards predicting output classes.
		Through experiments, we demonstrate the effectiveness of the proposed method in explaining the causal relationship between a concept and the predictive behaviour of a neural network as well as determining the interactions between different concepts through constructing a concept hierarchy.
	\end{abstract}
	
	\section{Introduction}
	Applications of Machine Learning (ML) and Artificial Intelligence (AI) as methods to help with automatic decision-making have grown to the extent that it has raised concerns about the trustworthiness of these methods.
	There have been recent policies and regulations all around the world that organizations should provide explanations for decisions made by their automated decision-making systems \cite{goodman2017european}.
	These concerns often exist whenever the problem at hand is not fully understood, explored or our knowledge of the problem is not complete.
	Knowing the reasoning of machine learning methods may also help with catching their unwanted behaviours by comparing the reasoning to experts' understanding of problems.
	An example of such unwanted behaviours is biases in decision-making.
	On the other hand, explanations can be used to extract the knowledge gained by these black boxes as well.
	Knowledge extraction can help with a better understanding of the AI view of the problem and the machine learning methods.
	
	Neural networks as one of the most promising forms of AI with high performance on classification problems like ImageNet challenge \cite{deng2009imagenet,russakovsky2015imagenet} have been criticized for their black box decision-making process.
	One of the most important questions asked about neural network's decisions is how a certain \emph{concept} influences the internal representation and eventually the output of a neural network.
	Here, a concept is a representation of a particular feature and is defined by a set of samples with that feature, against a random set \cite{Kim2018}.
	For example, for predicting the job title of a person from their image, this feature can be as simple as the colour of uniform (e.g. white or pink), background (e.g. office, ambulance or clinic), or presence of objects (e.g. stethoscope) in the image.
	
	Breaking down the decision, output or \emph{task class} of a given neural network into high-level humanly meaningful \emph{concepts} in the form of a post-hoc analysis has been an active area of research in the past few years.
	This process is done by inspecting the internal representations or \emph{activations} of the neural network.
	This approach may be called \emph{concept-based explanation} of neural networks.
	The training phase of the original network and the post-hoc explanation phase can be completely separate with separate datasets (one for task classes and one for concept classes) and be done by different persons independently.
	For instance, for predicting the job title of a person from their image, the goal is to determine whether having a clinic as an image's background affects the prediction of the job title to be a physician.
	Or how the presence of a stethoscope around the neck changes the prediction of the neural network.
	These methods do not require concept labels and task labels to be from the same set of samples.
	For example, the task of predicting if the job title of a person is physician can be represented by a set of physician images.
	But, the concept clinic may be represented by a set of clinic images and the concept stethoscope may be represented by a set of stethoscope images.
	
	\subsection{Nonlinear Concepts}
	
	Most concept-based methods often assume that a concept, if present in an activation space, should be linearly separable from non-concept samples \cite{Bau2017, Kim2018,goyal2019explaining, Zhou2018}.
	This assumption, however, does not necessarily hold, especially in the earlier layers of a network where the learned features are often not abstract enough to linearly separate concepts \cite{koh2020concept} or in later layers when concepts are mixed up to form higher-level concepts.
	This hinders such methods' ability in tracking the presence of a concept throughout the layers of the network.
	Another limitation comes from the assumption that the gradient of a section of a network with respect to the input is a good representation of that section \cite{Kim2018, Zhou2018, pfau2021robust}.
	Such a first-order approximation might be misleading.
	This issue has been extensively discussed for saliency maps ---which are also based on gradients approximations--- and have been proven misleading \cite{adebayo2018local,kindermans2017reliability}.
	
	In our method, we check the presence of a concept in a layer's activations by training a concept classifier ---a network with the same structure as the task classifier from that layer onward but trained to detect the concept.
	The accuracy of concept classification gives us a good understanding of the importance and possible influence of the concept on a task class.
	For instance, the colour of shoes might not be relevant in the job classification task and such information is likely to be discarded by the network after the first few layers.
	If in a particular layer, a concept cannot be detected by the concept classifier, it is safe to say that the network cannot recall the concept --i.e. the concept is forgotten (not necessarily universally but to the capacity and power of the given network).
	Such a conclusion can be made only if the concept classifier shares the same structure as the task classifier since the network structure is the upper limit for the extraction power of the network.
	Moreover, the concept classifier is initialized by the weights of the network under inspection.
	This initialization will reduce the number of concept samples required for training the concept classifier.
	This particular choice of the concept classifier's structure allows us to track the concept information across the original network's layers.

	\subsection{Causality}
	
	Another shortcoming of the previous methods is that most methods yield a score that captures the correlation between concepts and output and cannot give any further details about the nature of such a relation \cite{Bau2017, Kim2018, Zhou2018}.
	Following the above example about job title classification from images, the correlation between wearing a lab coat and being classified as a doctor cannot answer questions like ``do all images classified as doctor include lab coats?'' or ``do all people wearing lab coats are classified as doctors?''
	This problem is sometimes referred to as causality confusion.
	Note that the goal is not to investigate causal relationships in the training dataset.
	We aim to investigate the causal relations ``learned'' by a neural network.
	
	In a medical diagnosis setting one may ask questions like, 
	do all patients that are classified as having flu have fever symptoms (fever necessary for predicting flu)?
	Are all patients with fever classified as having flu (fever sufficient for predicting flu)?
	Are all patients classified as flu do not have a fever (absence of fever necessary for predicting flu)?
	Are all patients with fever not classified as flu (fever sufficient for negative of predicting flu)?
	What are the symptoms that if present together the predicted diagnosis will be flu?
	
	\subsection{Combination of Multiple Concepts}
	Most approaches are unable to analyze combinations of concepts, especially in cases where the behaviour is nonlinear.
	Linear assumption means that each concept attributes to a small part of the decision making and their influences add up.
	For example, consider a situation where symptoms (A and B), or (C and D) are enough for a diagnosis, but not mixed symptoms like (A and C), (B and C), (A and D) and (B and D).
	Here a linear model fails because it cannot detect (A and B) and (C and D) pairs without activating, for example, (A and C).
	Another example is when a task class is equivalent to the exclusive OR of two concepts, such concept relationships cannot be represented by linear models.
	To address this issue, we propose an approach for constructing a decision tree that can capture non-linear relations between concepts and provide a hierarchy of concepts.
	

	

	The main contributions of this paper can be summarized as follows: 
	\begin{itemize}
	\item We propose a general framework for quantifying causal relationships between a concept (or its negation) and task classes.
	While the previous approaches focused on the importance of a concept to a task class, we go further and introduce four measures to quantitatively determine whether the concept (or its negation) is necessary, sufficient, or irrelevant for a task class.
	
	\item While the mainstream previous work is based on linear models for representing concepts, we propose to use a non-linear model in the form of a neural network that its structure and initial weights are transferred from an appropriate segment of the original network.
	
	\item We propose an approach for constructing a hierarchy of concepts in the form of a decision tree.
	The decision tree can shed light on how various concepts interact inside a neural network towards predicting output classes. 
	
	\end{itemize}
	
	To avoid unnecessary assumptions like linear assumption or first-order approximation, we use a distribution sample set --i.e. a set of samples representing the distribution of the input data.
	This set is representative of the likely inputs of the network.
	Unlike the previous works in \cite{Bau2017, Zhou2018}, which are limited to specific network structures like convolutional layers, the proposed method can be applied to a wide range of network structures.
	A preliminary version of this work has appeared in \cite{zaeem2021cause}.
	Improvements compared to the preliminary version are two folds: first, we have expanded the proposed method to work on multiple concepts and propose a method for constructing a hierarchy of concepts in the form of a decision tree.
	Second, we have included new experiments as well as additional discussions.

	\section{Related Work} \label{Related Work}
	
	There have been several works on explaining intermediate activations of a neural network based on human-friendly concepts.
	Most notably, Kim et al. \cite{Kim2018} proposed a percentage measure, called TCAV score, to measure how much a concept interacts with the task classifier.
	TCAV works based on whether the gradient of the neural network is in the direction of the concept.
	The direction of the concept is defined as the direction orthogonal to the linear classification decision boundary between concept and non-concept samples.
	The TCAV score captures the correlation between the network output and the concept and lacks detailed information about the nature of the relationship.
	Moreover, it assumes that concepts can be represented linearly in the activations space, an assumption that does not necessarily hold \cite{koh2020concept}.
	They also represent a section of the network only by its gradient (first-order approximation), which might be misleading.
	Similar approaches have also been explored in Net2Vec \cite{fong2018net2vec} and Network dissection \cite{Bau2017} methods but they assume that the concepts are aligned with single neurons' activation.
	The derived works like RCAV (Robust Concept Activation Vectors) \cite{pfau2021robust} try to address the linearity problem, but the gradient approximation problem still persists.

	In another work, Interpretable Basis Decomposition for visual explanation (IBD) \cite{Zhou2018}, authors tried to explain the activations of a neural network by greedily decomposing it into some concept directions.
	They use the resulting decomposition as an explanation for the image classification task.
	One of the drawbacks of such an approach is its linear assumption which comes from the usage of linear decomposition of the gradient in the activation space.
	Using greedy methods can also potentially result in inaccurate and unstable results.
	Another limitation of the IBD method \cite{Zhou2018} is that it can only explain convolutional layers and therefore for neural networks that include dense layers, they have to modify the network.
	In their experiments, they have replaced each dense layer with a global average pooling layer and a linear layer.
	
	Concurrent to our work, Singla et. al. \cite{singla2021using} proposed that instead of linear decomposition, a decision tree can be used to explain neural network behaviour.
	However, they still use logistic regression for concept representation, which in nature is linear and has all the problems mentioned earlier.
	
	The linear assumption indicates that a concept in hidden layers corresponds to a vector and the representation of data in each layer is a vector space.
	Such methods assume that addition, subtraction, scalar product and inner product (as projecting an activation to a concept vector) operations in an activations space are always meaningful.
	The linear assumption is originated in feature visualization methods.
	Most feature visualization methods optimize for inputs that maximally activate certain neurons or directions.
	Early studies on neural network activation space tried to find samples that maximally activate a single neuron, and associate a concept to the neuron.
	In \cite{Szegedy2014} the authors argued that random linear combinations of neurons may also correspond to interpretable meaningful concepts.
	The general idea of using a linear classifier to check the information of intermediate layers originated in \cite{Alain2016}.
	They proposed to use linear probes -- trainable linear classifiers independent of the network -- to get an insight into the network representations.
	In contrast to what was mentioned in \cite{Szegedy2014}, in \cite{Bau2017, olah2017feature} the authors reported that the basis (each neuron) direction activation is more often corresponding to a meaningful concept than just random vectors.
	Still feature visualization methods, ignore the distribution of the input data which results in inputs that are not consistent with real samples.

	Linear interaction of concepts has been even less studied in feature visualization methods.
	In \cite{olah2017feature} the authors showed in some cases the addition of two concepts' activations will result in inputs with both concepts present.
	But they cast doubt on whether this finding is always true.
	Linear assumption lacks enough evidence to be considered reliable for being the basis of interpretability methods that try to gain the trust of humans and justify neural network decision-making.

	Some other methods have tried to automatically discover new concepts from neural networks, namely Automatic Concept-based Explanations (ACE) \cite{ghorbani2019towards} and Completeness-aware Concept-Based Explanations (CCE) \cite{yeh2020completeness} and others \cite{kamakshi2021pace,cheng2021game,ghandeharioun2021dissect}, rather than taking a concept as input.
	ACE tries to automatically extract concepts based on TCAV while CCE extracts concepts based on convolutional layers continuity.
	CCE also tries to avoid first-order approximation by measuring the importance of concepts using the Shapley score.
	Though these methods can help with cases that no principle exists for rational behaviour of the network, in many cases, the experts have a good principle about the problem at hand and the principle's concepts are predefined.
	So they want to check the consistency of the neural network with the existing principles.
	For instance, in the detection of a certain disease, the experts check all the related symptoms and are not interested in other concepts that a machine learning method might introduce.
	For example for the prediction of a patient having flu, medical experts know that fever is a symptom, and we want to know exactly what is the relation between the fever and being classified as having flu.

	Our work relates to CACE \cite{goyal2019explaining} in that, both try to address the shortcoming of TCAV \cite{Kim2018} by capturing causal expressions.
	The CACE method \cite{goyal2019explaining} measures the influence of concept by the difference of conditional expected values.
	This requires highly controllable datasets or very accurate generative models that may not be available in practice.
	Our method relates to works that define and train neural networks with concept-based explanations in mind \cite{koh2020concept, bahadori2020debiasing, chen2020concept, barbiero2021entropy}, though our method explains existing pretrained neural networks (post-hoc approach).
	
	Our work relates to \cite{wexler2019if} in that both use a specific visual method to examine the influence of different input features on the output of a machine learning model.
	But our method goes further and inspects the nature of the relationship and quantifies these visualizations.
	We also consider high-level human-friendly concepts instead of raw input features.

	In the next section, we will present the proposed framework which simultaneously addresses the linear assumption, first-order approximation and causality confusion issues discussed above.
	
	\section{Framework} 
	\subsection{Background}
	Logical expressions are usually expressed as a causality clause in the mathematical notation form of $A \Rightarrow B$.
	In this notation, phenomenon $A$ is the reason for the phenomenon $B$ and whenever $A$ happens, $B$ will follow.
	To understand the clause, both $A$ and $B$ should be understood.
	Meaning that both condition and consequence should be familiar for humans so that the clause can be understandable.
	
\begin{figure}
		\centering
		\includegraphics[width=0.4\textwidth]{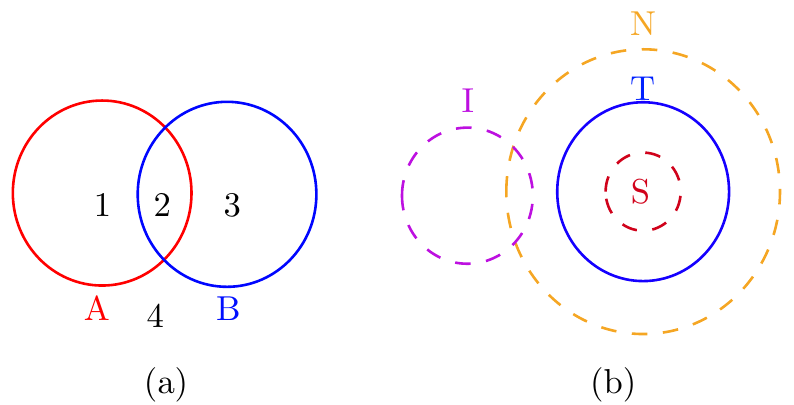}	
		\caption{ (a) There are four different subsets that might potentially be empty.
		Depending on which subset is empty, one of the four conditions will happen.
			(b) 3 relative positions of two sets.
			With respect to (T) Task class, (S) Sufficient condition, (N) Necessary condition, and (I) Negative necessary.}
		\label{fig:sets}
	\end{figure}

	In fundamental math, concepts are represented by sets.
	We use the same representation to visualize the relation between concept and task in a neural network.
	Being in a set means that the corresponding feature is present, and not being in the set means the feature is not present.
	Two arbitrary sets are usually demonstrated by a Venn diagram (see Figure \ref{fig:sets}(a)).
	
	There can be several possible relations between the two sets.
	Each of these relations can also be represented as a causality clause.
	\begin{itemize}
		\item Necessary: $T \subseteq C$ ($T \Rightarrow C$).
		\item Sufficient: $C \subseteq T$ ($C \Rightarrow T$), reverse of necessary.
		\item Negative Necessary: $C \cap T = \emptyset$ meaning $C$ and $T$ are inconsistent ($C \Rightarrow \neg T$ or $T \Rightarrow \neg C$).
		\item Negative Sufficient: $C \cup T = M$ meaning either $C$ or $T$ or both should happen ($\neg C \Rightarrow T$ or $\neg T \Rightarrow C$). ($M$ is the set that contains all the elements).
	\end{itemize}
	

	\subsection{Determining Relevant Concepts}
\begin{figure*}
		\centering
		\includegraphics[width=\textwidth]{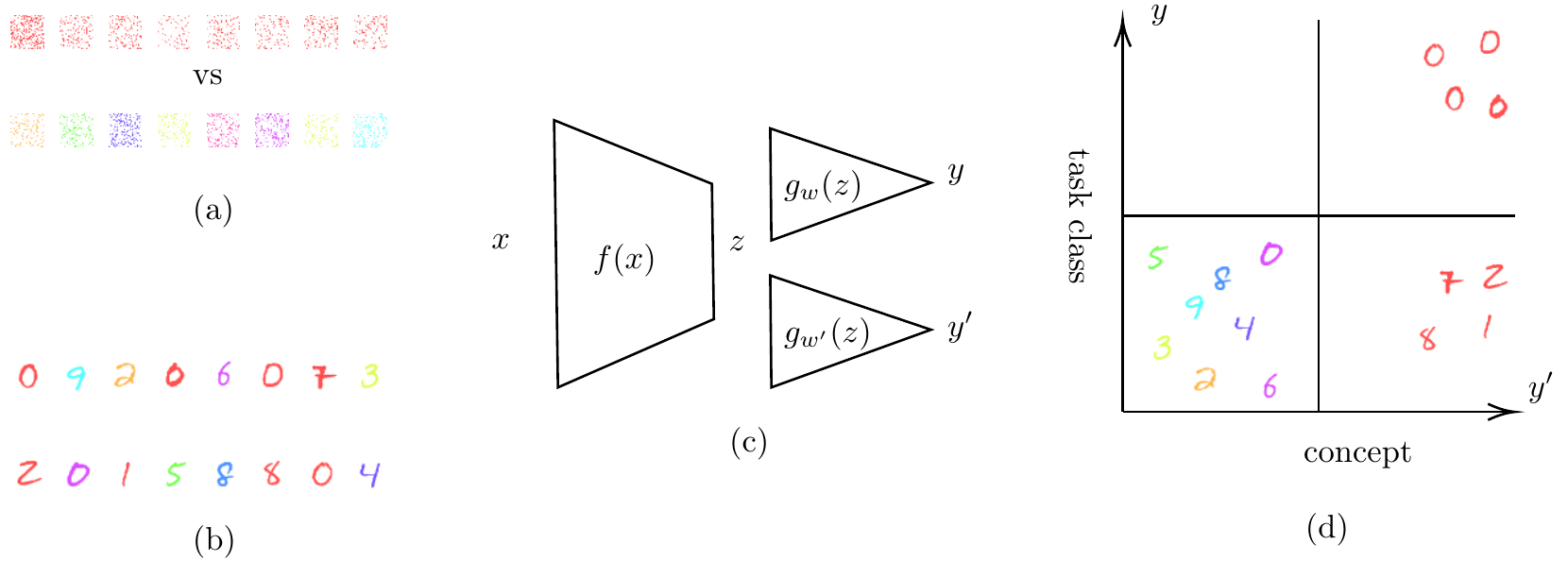}
		\caption{
			(a) Concepts set for the concept red. (b) Distribution sample set. (c) Illustration of original network as $f(g_w(z))$, task class network $g_w(z)$ and the concept network $g_{w'}(z)$. (d) Analysis of the relationship between the task class and concept showing the concept is a necessary condition for the task class.		
			Here the task class was defined as the classification of digits where each class has a unique colour.
		}
		\label{fig:scatter}
	\end{figure*}

	In our method, we base the explanations on a certain layer's activations and explain whether and how a concept interacts with a task class based on the activations of the layer.
	We break up the neural network into two sections, the section before the hidden layer (denoted by $f(x)$) and the section after the hidden layer (denoted by $g_w(z)$).
	$w$ denoted the trainable parameters of the second section and the whole network can be expressed as $g_w(f(x))$ (see Figure \ref{fig:scatter}(c)).
	
	As shown in Figure \ref{fig:scatter}, we only need two sets of samples for our analysis.
	(1) Concept set labelled on the concept information only (Figure \ref{fig:scatter}(a)).
	(2) Distribution sample set, without any labelling (Figure \ref{fig:scatter}(b)).
	Note that access to the training data of the original tasks is not required.
	
	For the sake of explaining the proposed method, let us consider a neural network trained on colour-coded hand-written digits.
	In the training set, a unique colour was assigned to each class, and samples within each class were coloured accordingly.
	For instance, all 0's in the training set are red, and all 1's are blue.
	One would expect the network decision to be influenced by the colour as well as the digit itself; but the challenge is how to measure this influence.
	We aim to determine, how a concept, e.g. red, influences the decision-making of this neural network.
	The first step in our analysis is to check whether a concept is present in a layer.
	In other words, can a classifier trained on the activations of that layer achieve an acceptable concept detection accuracy.
	We check if the second section of the network has adequate power and capacity to distinguish the concept set (colour red) from random none-concept samples in that layer.
	Since detecting a concept is a binary classification, the number of output neurons is adjusted accordingly.
	
	For representing a concept, a set of positive and negative concept examples are used, in our case red samples against other colours (Figure \ref{fig:scatter}(a)).
	Then the concept classifier ($g_{w'}(z)$) with the structure of the second section of the neural network ($g_w(z)$) is trained to distinguish the concept from non-concept samples.
	As a result, we will have a network with identical structures as the task classification but different parameters.
	$w$ denotes the parameters of the original network trained for task classification (digit classification) and $w'$ denotes the parameters learned to distinguish a concept from non-concept samples (red vs. other colours).
	$g_w(z)$ is the task classifier, whereas $g_{w'}(z)$ is the concept classifier.
	For learning $g_{w'}$, we initialize it with the weights of the task classifier i.e. $w$.
	Note that the parameters of the first section ($f(x)$) do not change while $w'$ is learned.

	Two observations support our choice for using the same structure for the detection of concepts.
	First, if the concept is present and the network is using it, the network must have extracted the concept information using its structure.
	Consequently, the network structure should be able to detect that concept.
	Second, if the concept is not detectable by the network structure, there is no way for the concept to have influenced the task class predictions.
	Although concepts like colour can be easily learned by a much simpler network structure, more complex concepts like the presence of objects (a stethoscope) might not be as simple to detect.
	Moreover, using the same structure allows us to show the absence of a concept in a layer in the sense that it cannot be detected using the second part of the network structure (from $z$ to output) and consequently conclude that the concept could not have had any influence on the task class.
	Note that we do not need to determine the absence of a concept globally.
	If in a global sense, a concept is present in a layer but the second part of the network structure does not have enough capacity to learn to detect it, for the purpose of determining if the concept has any influence on the task class, we can consider the concept as being absent.
	Another advantage of using the same structure is that we can initialize the concept networks using the weights of the original network.
	Therefore, the concept networks will be pre-trained on task classification and hence indirectly on the concepts involved in the task classification.
	Moreover, the structural coherence of the network is kept intact.
	In other words, the limitations, powers and local behaviour of the network (as initial parameters) are considered in the detection of concepts, by keeping for example convolutional activations as convolution representations (with the spatial information preserved).
	
    \subsection{Quantifying Causal Relationships between Concepts and Task Classes}
\begin{figure}
		\centering
		\includegraphics[width=0.3\textwidth]{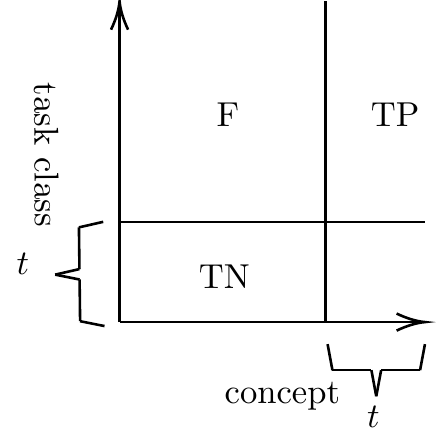}	
		\caption{The process of constructing quantification curve for necessary score $T \Rightarrow C$.
		In this process we verify if the task class is true with probability at least $t$ then the concept is true with probability of at least $1-t$.
		The highest value is achieved when there is no counter-example ($F$) samples.}
		\label{fig:multithresholding}
	\end{figure}

	Checking if a set is a subset of another, can be easily done by checking the definition.
	Since we cannot sample every possible instance in the input space, we only check the relationship on a distribution sample set (Figure \ref{fig:scatter}(b)).
	Note that this sample set is chosen randomly and it is not specifically selected like prototyping methods.
	As we mentioned earlier, the distribution sample set, the samples used for causal analysis, does not have any labels.
	Labels are not needed because the distribution sample set will be used only for quantifying the causal relationships between a concept and task classes after the concept network is trained using the concept set (more on this will follow).
	The set $T$ is a subset of set $C$ if every sample in $T$ is also in $C$, which is equivalent to $C$ being a necessary condition for $T$ or $T \Rightarrow C$.
	Checking the negation of this definition is much easier (just checking that no counter-example exists).
	For this purpose, a scatter plot is generated by evaluating the task classifier and concept classifier on the distribution sample set --see Figure \ref{fig:scatter}(d).
	Each point on the scatter plot is a sample from the distribution sample set.
	A counter-example, in this case, is a sample in $C$ and not in $T$ (e.g. a sampled classified as 0 and not red).
	So the clause correctness corresponds to the case where the top left corner of the scatter plot is empty -- equivalently no counter-examples found in our distribution sample set (Figure \ref{fig:scatter}(d)).
	Note that the points of the scatter plots are outputs of the task classifier ($g_w(z)$) and the concept classifier ($g_{w'}(z)$) based on the activations at an intermediate layer and the predictions are not necessarily close to true labels.
	We want to measure the relationship between the predictions of the concept and the task class networks regardless of the true labels.
	Likewise, other corners of the scatter plot being empty corresponds to other types of relationships between concept and task class.	
	Since the concept classifier and task class are represented by soft decisions (output probabilities), we propose a method to quantify the absence of counter-examples based on a modified F1 score.
	We prefer the F1 score over accuracy because it can better handle imbalanced classes.
	F1 score is defined as the harmonic mean of precision and recall.
	Precision and recall are determined when the predicted values are transformed to binary using a cutoff threshold.
	In a similar fashion, we transform the prediction of the concept and task class networks to binary using a cutoff threshold $t$ (see Figure \ref{fig:multithresholding}).
	Similar to the notion of the ROC curve, we will vary the threshold $t$ and plot the F1 score (more precisely an adapted F1 score) for different values of the threshold.
	
	To understand Figure \ref{fig:multithresholding} consider the process of constructing quantification curve for necessary score i.e. the logical expression $T \Rightarrow C$.
	It is equivalent to $\neg T \vee C$ which implies that for the logical expression $T \Rightarrow C$ to be true, $\neg T$ or $C$ has to be true.
	Accordingly, a counter example for $T \Rightarrow C$ happens when $\neg (T \Rightarrow C) = T \wedge \neg C$ is true.
	$\neg C$ can be calculated as $1-C$ since $C$ is a probability.
	$\neg C$ is greater than a threshold $t$ is equivalent to  $C$ being less than $1-t$.
	To reflect this, we compare the task class probability against the threshold $t$ and the concept probability against the threshold $1-t$.
	Therefore, the logical expression $T \Rightarrow C$ is true for any sample that is outside the region $F$.
	That is, when task class probability is true with a probability of at least $t$, then the concept is also true with a probability of at least $1-t$.
	
	
	F1 score is defined as follows:
	\begin{equation}
		F1 = \frac{2 P R}{P+R}\:.
	\end{equation}
	where $P$ and $R$ are adapted precision and recall defined as:
	\begin{equation}
		P = \frac{TP}{TP+F}\:,
	\end{equation}
	\begin{equation}
		R = \frac{TN}{TN+F}\:.
	\end{equation}
	
	TP, TN and F denote the number of samples in the corresponding regions of Figure \ref{fig:multithresholding}.
	
	Based on the introduced parameters, for each threshold $t$, an F1 score can be calculated.
	The strength of a necessary relationship is then calculated as the area under the F1-versus-threshold curve.
	Intuitively the strongest necessary relationship happens when we have larger F1 scores for smaller values of $t$.
	That is, even a weak probability of belonging to the task class would require the concept to be present with a high probability.
	
	For simplicity here, we assumed that the threshold for $C$ and $\neg T$ are the same, but in general, these thresholds can be considered as threshold $t_C$ for $C$ and threshold $t_T$ for $T$.
	In that case, the quantitative curve will be a 3D surface (the F1 score vs. $t_C$ and $t_T$) and the volume under the curve should be used as the measure of the strength of the relationship.
	
	
	
	Each of the four relationships can logically be converted to an OR ($\vee$) expression and evaluated in the same manner we evaluated the necessary relationship.
	A simpler way is to logically convert the other three cases to a necessary condition and quantify them with the mentioned process.
	For instance $C$ being sufficient for $T$ is equivalent to $T$ being necessary for $C$.
	The negation of concepts and tasks ($\neg C$) can be calculated by subtracting them from one, i.e. ($1-C$), so for the negative necessary measure, we do a similar calculation but with one minus the values of the concept probabilities (more on this in Section \ref{Analysis of Causal Relationships}).
	
	For each of our experiments, we create four quantification curves based on the scatter plot.
	Each quantification curve shows the F1 score against different thresholds.
	We further use the area under the curve (AUC) to summarize each curve into a real-valued score between 0 and 1.
	An AUC value higher than $0.5$ would indicate the existence of evidence for the corresponding relationship.
	For example, for the concept stereoscope and the task class physician, we expect a large necessary AUC.
    An AUC value less than $0.5$ indicates that there is evidence against the corresponding relationship.
    For example, for the concept wrench and the task class physician, the necessary AUC is expected to be closer to $0$.
    An AUC value around $0.5$ would indicate there is no measurable relationship.
    For example, for the concept skin colour and the task class physician, we expect to see a necessary AUC of $0.5$.
    In general, when the two variables $T$ and $C$ are probabilistically independent and have uniform distributions, it can be proved that the AUC will be equal to 0.5 as follows:
	\begin{equation}
		P = \frac{TP}{TP+F} = P(C>1-t|T>t) = P(C>1-t) = t
	\end{equation}
	\begin{equation}
		R = \frac{TN}{TN+F} = P(T<t|C<1-t) = P(T<t) = t
	\end{equation}
	and then the $F1$ score will be
	\begin{equation}
		F1 = \frac{2 P R}{P+R} = t
	\end{equation}
	and the AUC we will be
	\begin{equation}
	    AUC = \int_{t=0}^1 F1\: dt = \int_{t=0}^1 t \: dt = 0.5 .
	\end{equation}
	
	\subsection{Concept Hierarchy}
	Now that we have determined the presence of concepts in the embedding space, we can move on to quantifying the relationships between concepts and construct a hierarchy of the relevant concepts.
	To this end, we select the concepts that are detected in the embedding space -- i.e. the accuracy of their corresponding concept classifier is high, 
	and train a decision tree that tries to estimate the task classifier as an interpretable function of the detectable concepts.
	At each node of the decision tree, we split based on a concept. 
	The decision tree will show the importance of different concepts for each class and their relative relationships.
	Note that this decision tree is estimating the task classifier and not the ground truth and its behaviour will estimate the network behaviour.
	
	This decision tree will discover what combination of concepts is considered for a task class.
	Since the distribution sample set represents the real sample distribution, we can describe, how a sample should look in terms of the concepts for the network to classify it as a particular class.
	A decision tree can be constructed in two ways:
	A single decision tree can be trained to decide about all task classes based using all the present concepts.
	Alternatively, we can train a decision tree per task class and determine the hierarchy of related concepts per task class.
	The former provides a higher-level understanding of the multi-class problem as a whole.
	The latter allows us to isolate and quantify related concepts for each task class more effectively and results in a smaller tree.
	In our experiments, we examined both approaches.
	
	The combination of concept classifier networks and the decision tree can be considered as an approximation of the original network.
	The accuracy of the decision tree combined with the accuracy of the concepts used for constructing it determine how faithful the combined model is to the original network.
	Assuming the concepts are simpler than task classes, it is much easier to verify the accuracy and correctness of concept classifiers and then explain the original network through the decision tree.
	Note that the concept classifiers and the decision trees are for explainability purposes, and it is the original network that does the classification task.
	
	Low faithfulness accuracy (accuracy based on network outputs and not the true labels) of the resulting decision tree may suggest that the considered concepts are not all the information that the original network is using, and introducing other concepts can potentially make the decision tree more faithfulness to the task classifier.
	Regardless, the proposed method can quantify the relation between concepts and task classes and also the interactions between the provided concepts even if the provided concepts do not fully represent all the information the original neural network uses.
		
	\begin{figure}
		\centering
		\includegraphics[width=0.3\textwidth]{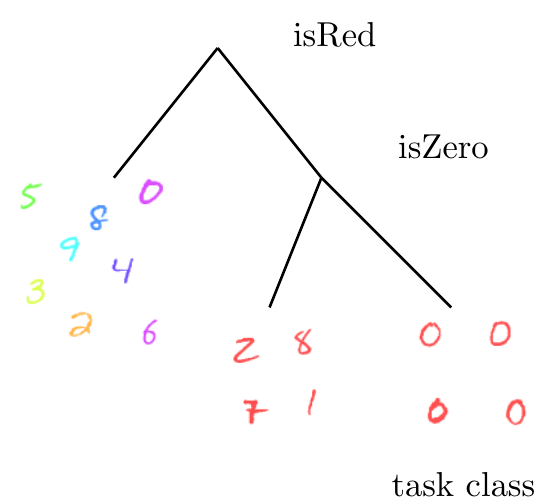}	
		\caption{
			Analysis of the relationship between the task class and concepts using a decision tree.
			The decision tree shows that both concepts of isRed and isZero are necessary for the task class ``Red Zero''.
		}
		\label{fig:scatterTree}
	\end{figure}

	
	As a byproduct of the resulting decision tree, each leaf can be considered as a new compound concept discovered by the decision tree.
	A compound concept consists of all the concepts along the path from the root to a leaf.
	The newly discovered compound concepts can go through the same process as the original concepts and the four measures discussed earlier can be calculated for them.
	To this end, a dataset consisting of the compound concept (i.e. intersection of the related original concept sets) versus some random samples can be constructed.

	
	In the next section, we will demonstrate the proposed methods over several experiments.
	
	\section{Experiments and Results}
	In this section, we explore the application of the proposed methods in the evaluation of the relationship of neural network task classes and concepts in a controlled setting, a real-world setting and a medical application.
	The experiment in the controlled setting explains a neural network with Alexnet structure and the experiment in a real-world setting uses a pretrained Resnet18.
	We compare our results with TCAV \cite{Kim2018} and IBD \cite{Zhou2018} methods as they are the most related work to the proposed method.
	We have constructed two controlled datasets to simulate different scenarios where concepts have certain positive or negative relationships with task classes.
	
	\subsection{Coloured MNIST Dataset}
\begin{figure}[!h]
		\centering
		\includegraphics[width=0.4\textwidth]{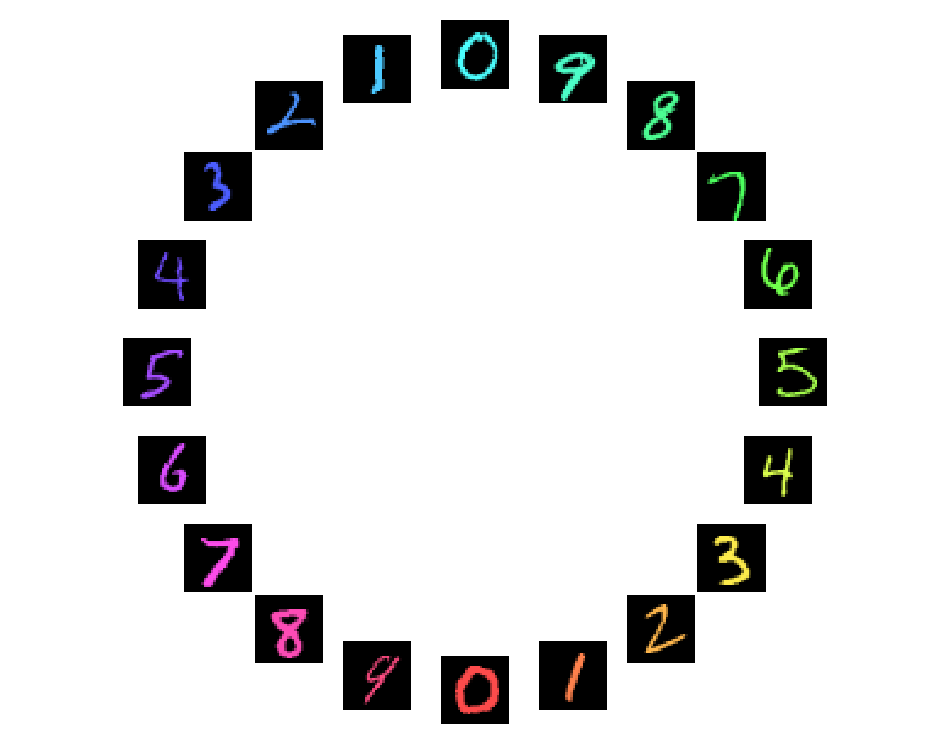}
		\caption{Illustration of how the two hint colours are chosen for each class in the ColorDataset1.
			Each class is associated with two different colours that have the highest distance in the colour space.}
		\label{fig:MNISTcoloredSamples}
	\end{figure}

	MNIST is an image recognition benchmark dataset consisting of ten classes of handwritten digit images \cite{lecun1998mnist}.
	We modify this dataset to add useful or useless additional hints (as colour) to the neural network that will be trained on this data.
	In this way, we will train various neural networks with desired characteristics in terms of using certain concepts. 
	Each of the ten task classes may correspond with multiple colours.
	The colour of each digit in a task class is chosen randomly from a set of two colours.
	We simulate two scenarios: 1) each digit is coloured using its own set of colours, --i.e. the colour concept and task class are fully correlated.
	We call this dataset ColorDataset1.
	2) all colourings are random (from the same set) --i.e. no relation between the colour concept and task class.
	We call this dataset ColorDataset2. 
	We choose the colours of each class (its colour set) in a way that no two classes (colour sets) in the colour space can be linearly separated, see Figure \ref{fig:MNISTcoloredSamples}.
	This is to simulate the situation where concepts are not linearly separable analogous to an XOR problem which cannot be solved by a linear classifier.
	For generating the concept samples (for training the concept classifier), we shuffle pixels of images (to wipe out the image information) and then add colours to the foreground pixels.
	Each colour set (two colours) is considered a concept.
	
	\subsection{Images with Captions Dataset}
\begin{figure}[!h]
		\centering
		\includegraphics[width=0.4\textwidth]{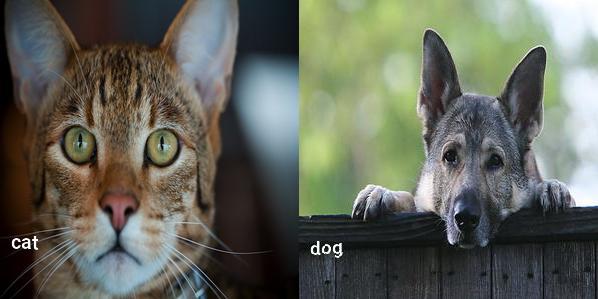}
		\caption{Two samples from CaptionDataset1.}
		\label{fig:imagenetSamples}
	\end{figure}

	In this dataset, we add a hint caption to two classes of the Imagenet dataset \cite{deng2009imagenet,russakovsky2015imagenet}, namely class dog and class cat.
	The hint is added as white text on the image (by changing the pixels of the image).
	Consequently, some pixels of each sample carry extra information about the task class.
	We consider two scenarios: 1) The caption always reads the same as the image (the word cat for cat images, and dog for dog images).
	We call this dataset CaptionDataset1.
	2) The caption is always a random word and hence does not include any information about the classification task (dogs vs. cats).
	We call this dataset CaptionDataset2.
	The captions have random rotation and scaling associated with them.
	Figure \ref{fig:imagenetSamples} shows two samples of the CaptionDataset1 images.
	
	For generating the concept samples (caption concept), we shuffled pixels of images (to wipe out the image information) and then added a caption to the resulting shuffled image.
	
	\subsection{Analysis of Causal Relationships between Concepts and Task Classes} \label{Analysis of Causal Relationships}
\begin{figure}[!h]
		\centering
		\includegraphics[width=0.5\textwidth]{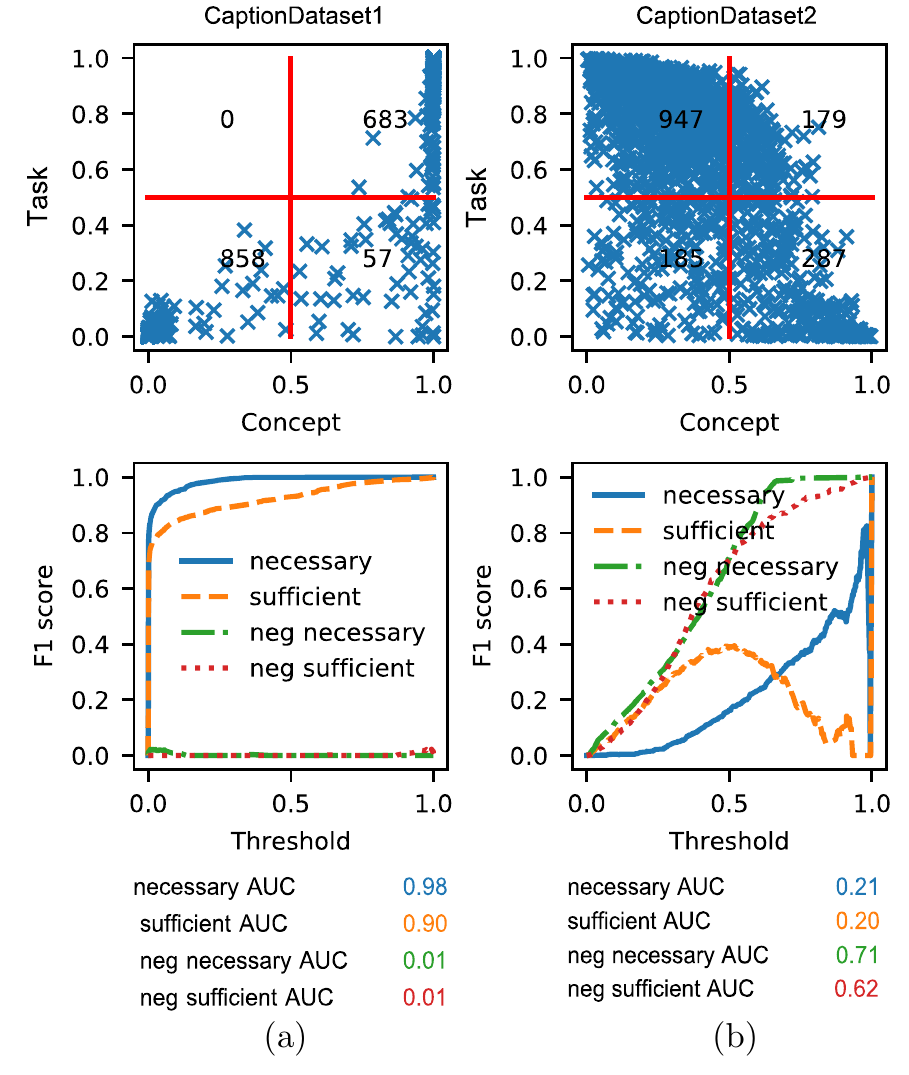}
		\caption{Comparison of two networks with the same structure, (a) has been trained on CaptionDataset1, (b) has been trained on CaptionDataset2.
			The AUC of one for the necessary curve shows that the concept is necessary for the task class.
			The corresponding AUC values are mentioned at the bottom.
			These results are extracted from the last hidden layer of a neural network with an Alexnet structure.
		}
		\label{fig:effectivenessCaption}
	\end{figure}

	In this experiment, we show the effectiveness of the proposed method in detecting the causal relationship between a concept and task classes of neural networks.
	We train an Alexnet on each of our datasets.
	The results for CaptionDataset1 and CaptionDataset2 are shown in Figure \ref{fig:effectivenessCaption} (fifth layer).
	Similar results were obtained on the ColorDataset1 and ColorDataset2.
	On the left side (a), it can be seen that the concept (caption dog) was detected to have a 98\% necessary relationship with the class dog.
	All samples predicted by $g_w$ as the dog class (above the red horizontal line) are predicted by $g_{w'}$ to have the caption concept (i.e. they are on the right of the red vertical line).
	Here, the red lines show a threshold of $t=0.5$.
	Plotting the modified F1 score for different values $t$ gives the necessary quantification curve (blue), which as expected has a high AUC (0.98).
	The sufficient quantification curve (orange) can be obtained from a scatter plot of $C$ vs. $T$ (instead of $T$ vs. $C$) because $C$ being sufficient for $T$ is equivalent to $T$ being necessary for $C$.
	The negative necessary quantification curve (green), can be obtained from the scatter plot of $T$ vs. $1-C$ and the negative sufficient quantification curve (red), can be obtained from the scatter plot of $1-C$ vs. $T$.
	The results for CaptionDataset2 are shown on Figure \ref{fig:effectivenessCaption} (b).
	It can be seen that, as we expected, there is no tangible relationship between the dog class and the caption concept.
	This confirms that the proposed method detects the causal relationship between the caption concept and the task classification.

	Now that we have established that the method can detect the usefulness of the hints, from now on we only use the CaptionDataset1 and ColorDataset1.
	
	\subsection{Informativeness of the Proposed Relationship Measures}
\begin{figure}[!h]
		\centering
		\includegraphics[width=0.5\textwidth]{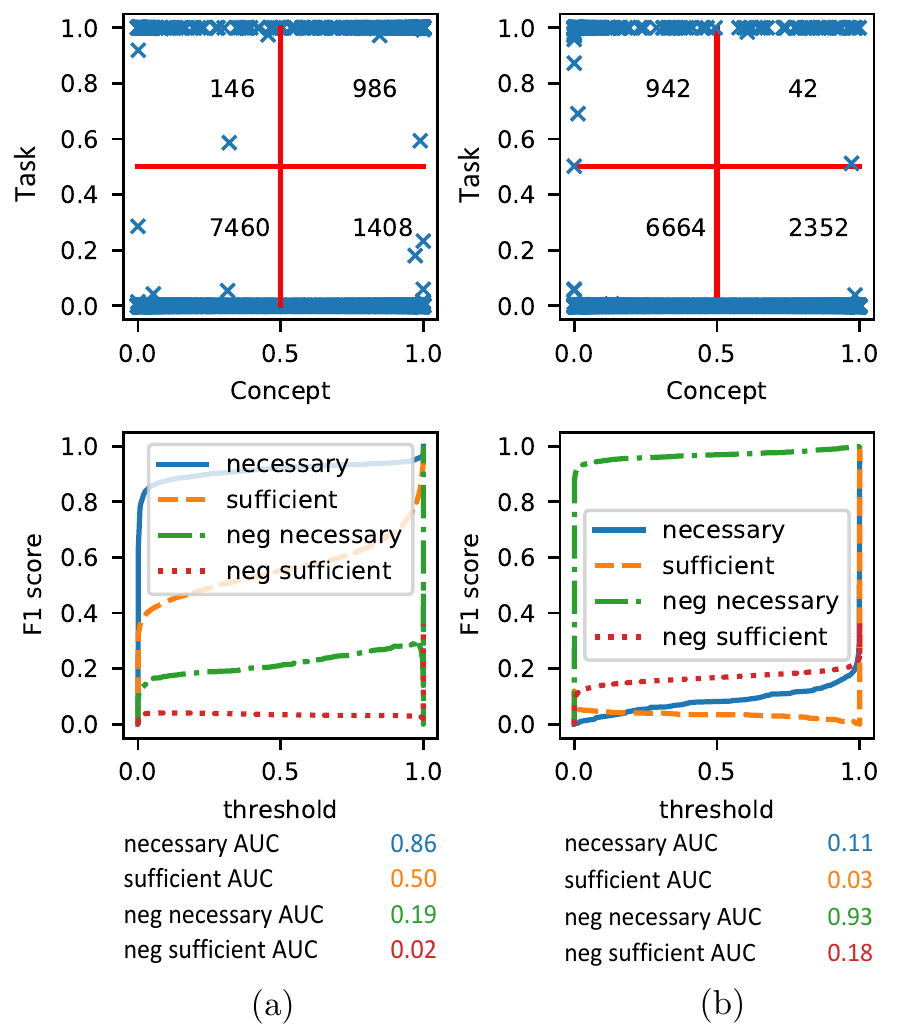}
		\caption{
			The causal relationship for the concept of the colour set of class one in the ColoredDataset1. (a) this concept is necessary for class one. (b) this concept is a negative necessary for all other classes.
			For example results for class zero are shown on the right.
			}
		\label{fig:informativeness}
	\end{figure}

	In this experiment, we show how the proposed relationship measures can reveal more than just a correlation between a concept and a task class.
	We examine the third layer of an Alexnet with ten classes, trained on our ColorDataset1 to check the causal relationships between different target classes and a particular concept.
	In particular, we inspect the concept of the colour set associated with the class of handwritten digit one, which is a \DigitOneBlue{shade of blue} and \DigitOneRed{a shade of red} as shown in Figure \ref{fig:MNISTcoloredSamples}.
	By design, this concept is a necessary condition for class 1 and negative necessary for all other classes.
    In other words, for class 1, it is necessary to have the concept (if the colours are not present, the task is not classified as class 1), but for other classes, like class 0, it is necessary not to have this concept (if the colours are present, the output will not be class 0).
    First, we quantitatively measure if this concept is a necessary condition for class 1.
    The results are shown in Figure \ref{fig:informativeness}(a).
    It can be seen that the proposed method successfully calculated a large necessary score i.e. AUC=0.86.
    Second, we quantitatively measure if this concept is a negative necessary for class 0. 
    The results are shown in Figure \ref{fig:informativeness}(b). 
    It can be seen that the proposed method successfully calculated a large negative necessary score i.e. AUC=0.93.

        
		

	\subsection{Comparison with a Linear Concept Classifier}
\begin{figure}
		\centering
		\includegraphics[width=0.5\textwidth]{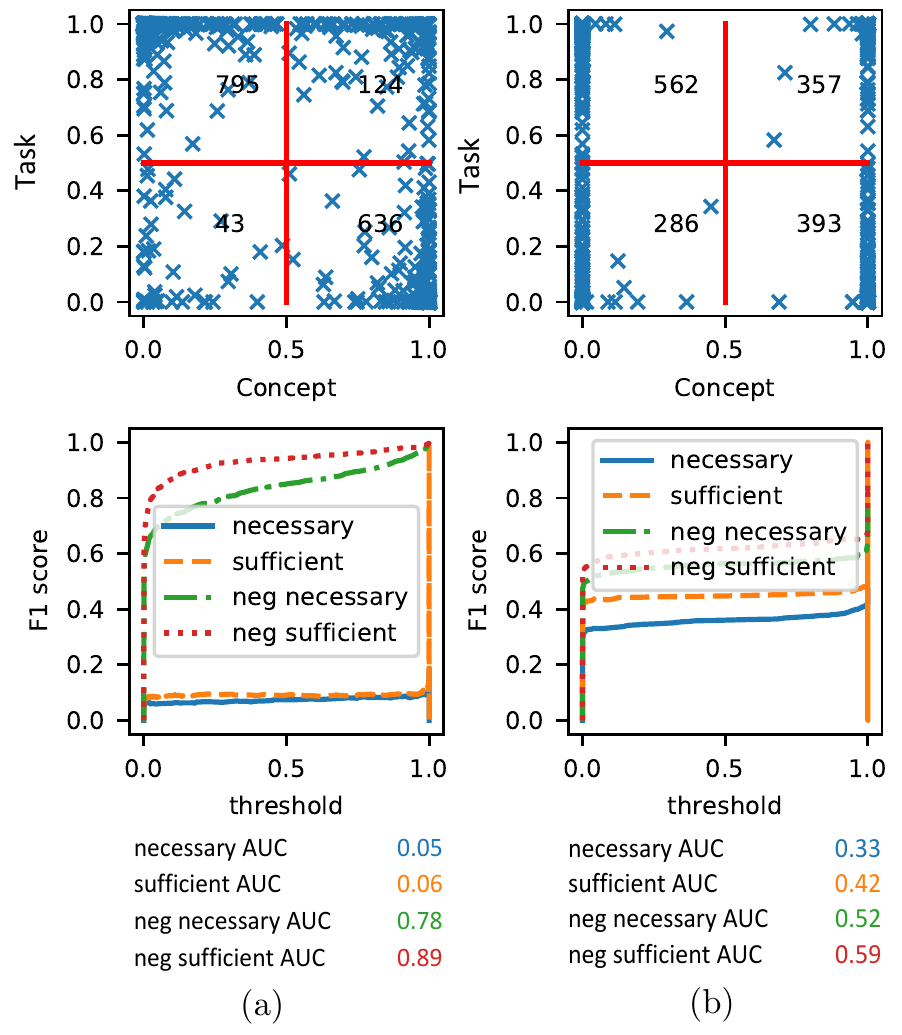}	
		\caption{Comparison with a linear classifier.
		We consider the concept of caption cat and its relationship to the task class dog. (a) Results of using the proposed concept classifier $g_{w'}(z)$. (b) Results of using a linear concept classifier.
		By design, the concept is negative necessary and negative sufficient for the task class dog.
		Unlike the linear model, the proposed method has easily detected these relationships.
		}
		\label{fig:linearComparison}
	\end{figure}

	As discussed in Section \ref{Related Work}, many methods use linear classifiers to extract concept information from a middle layer of neural networks \cite{Kim2018,Zhou2018}.
	In this experiment, we check the implications of using a linear classifier to detect concepts in network activations.
	We check the effectiveness of such an assumption in detecting the caption concept in an earlier layer (third layer) of Alexnet trained on our CaptionDataset1.
	By design, caption dog is both necessary and sufficient for the task class dog --i.e. all dog images and only dog images have the dog caption. 
	Also, caption cat is a negative necessary and negative sufficient for the task class dog --i.e. it is necessary and sufficient not to have a caption cat to be class dog. 
	In Figure \ref{fig:linearComparison} we have quantified these relationships using two types of concept networks.  
    On the right, we show the results of using a linear concept classifier i.e. we replaced $g_{w'}(z)$ with a linear classifier.
    On the left, we show the results of using a non-linear concept classifier, in particular, we use $g_{w'}(z)$ as a concept classifier.
	It can be seen that the proposed method has successfully quantified this relationship where the AUC scores for the negative necessary and negative sufficient conditions are high (78\% and 89\% respectively).
	But with a linear classifier, the four relationship AUC scores have almost similar values, and the designed relationship cannot be detected.
	These findings are consistent with what was reported in \cite{koh2020concept}.
	
	This experiment shows that the linear classifiers have failed to extract concept information in earlier network layers where activations usually represent low-level features.
	This phenomenon might be observed even at later layers of a neural when the concept is merged with other concepts to create a higher-level concept.
	(see Figure \ref{fig:sortingProblem1}).

	\subsection{Comparison with TCAV} \label{Comparison with TCAV} 
\begin{figure}[!h]
		\centering
		\includegraphics[width=0.5\textwidth]{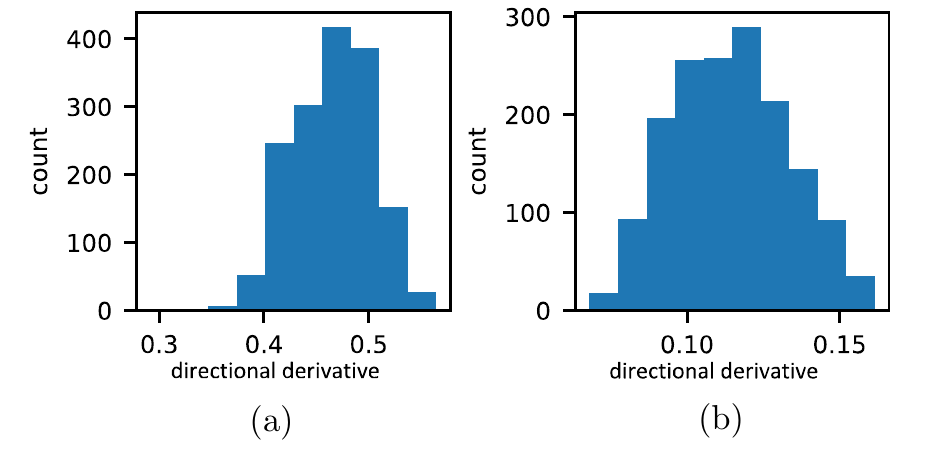}
		
		\caption{ Histogram of directional derivatives for the caption experiment discussed in Section \ref{Comparison with TCAV} using (a) a linear concept classifier, and (b) a nonlinear concept classifier.
			}
		\label{fig:sensetivityComparison}
	\end{figure}

	In many methods, the directional derivation of a classifier with respect to activations or input is considered as a good representation of the model's local behaviour (first-order approximation) \cite{Kim2018,Zhou2018}.
	As it was shown in TCAV \cite{Kim2018}, the directional derivative can be calculated by the dot product of the concept gradient and the task gradient.
	These methods use the dot product of a concept classifier gradient and a task classifier gradient as an agreement score evaluated on different class samples (linear assumption).
	Here we demonstrate an example of where this score is not accurate.
	We examine the relationship between the concept caption cat and the task class dog in the fifth layer of an Alexnet trained on CaptionDataset1.	
	By design, these two pieces of information are inconsistent in the training data, --i.e. no training sample from class dog has caption cat. 
	The distribution sample set consists of dog images with either caption dog or cat, and cat images with either caption dog or cat --i.e. all four possible combinations of cat and dog images with cat and dog captions. 
	Therefore, for half of the distribution sample set the image class and the added caption match and for the other half, images have a wrong caption.
	Figure \ref{fig:sensetivityComparison} (a) shows the distribution of directional derivatives between the concept and class using a linear model (similar to TCAV).
	Though the concept and class are by design inconsistent, the resulting directional derivatives are positive on all samples of the distribution sample set, showing directional derivatives are not a reliable explanation.
	Similar results observed using a non-linear classifier $g_{w'}$ (see Figure \ref{fig:sensetivityComparison} (b)).
	Note that the proposed method successfully captured the correct relationship (negative necessary) with an AUC of 96\%.

	\subsection{Comparision with the IBD method}
	Since the IBD method \cite{Zhou2018} is limited to convolutional networks, in this experiment we modify our method to be comparable to IBD.
	We examine the last hidden layer of a Resnet18 trained on the Places365 dataset \cite{zhou2017places} -- a dataset where each class is a place.
	This network was the benchmark of the IBD method.
	We use the same set of concept classifiers they trained (with the parameters they provided).
	We use 10,000 samples from the places365 validation set without their labels as the distribution sample set.
	Our concepts come from the same dataset the IBD method used as their benchmark, Broden \cite{Bau2017} -- a dataset with segmentation annotations.	
	The pixel-wise annotation of the Broden dataset can improve the training process if the output of the network is modified so it can reflect the spatial information preserved in the annotations.
	This technique improves the concept classifier accuracy and provides localized behaviour of the concepts.
	Such labels can be used to visualize as heat maps where the network detects specific concepts like the heatmaps in \cite{Zhou2018}.
	For the sake of comparison, we use only use the concepts originally used in the IBD benchmark.
	
	This experiment is designed to find the most important concepts for classifying each class of the Places365 dataset.
	For each concept, a concept classifier is trained, and then each task class (a class from Places365) is examined against each concept.
	The necessary scores of concepts for each task class are sorted and the highest values are reported as the most necessary concepts for the class.
	The most necessary concepts are then compared against the concepts recommended by IBD which are based on the decomposition of the decisions into concept space.
	The top seven are reported for both methods in Tables \ref{tab:explain}).
	The concepts are from left to right in decreasing importance.
	Three human annotators were asked to label the concepts as either relevant or irrelevant to the corresponding class.
	A concept is labelled as irrelevant if the majority of the annotators have found it to be irrelevant.	
	Irrelevant concepts are shown with a strikethrough text in the Table \ref{tab:explain}).
	It can be seen that the proposed method has identified more relevant concepts than the IBD method.
	The proposed method has assigned more reasonable values of necessary scores to the concepts compared to what IBD calculates based on its decomposition process.
	For instance, for the task class topiary garden, IBD suggested the concepts tail and sheep (among five others) which are irrelevant to the class of topiary garden.
	On the other hand, our method suggested plant and tree which are quite relevant concepts to the topiary garden class.
	For the soccer field class, our method suggested grass, pitch, grandstand, court, person, post and goal which are all relevant.
	But IBD suggested pitch, field, cage, ice rink, tennis court, grass, and telephone booth where cage, ice rink, tennis court, and telephone booth are irrelevant to the soccer field class (see Table \ref{tab:explain}).
	
	\begin{table*}[!h]
		\centering
		\begin{tabular}{|c|c|c|c|c|c|c|c|}
			\multicolumn{8}{l}{\textbf{Class: topiary\_garden}} \\ \hline
			Proposed	&plant	&hedge	&tree	&brush	&flower	&bush	&sculpture \\ \hline
			IBD	&hedge	&brush	&\cancel{tail}	&palm	&flower	&\cancel{sheep}	&sculpture \\ \hline
			\multicolumn{8}{c}{}\\
			
			\multicolumn{8}{l}{\textbf{Class: crosswalk}}\\ \hline
			Proposed	&crosswalk	&road	&sidewalk	&post	&\cancel{container}	&streetlight	&traffic light\\ \hline
			IBD	&crosswalk	&minibike	&pole	&rim	&\cancel{porch}	&central reservation	&van\\ \hline
			\multicolumn{8}{c}{}\\
			
			\multicolumn{8}{l}{\textbf{Class: living\_room}}\\ \hline
			Proposed	&armchair	&sofa	&back	&cushion	&back pillow	&coffee table	&ottoman\\ \hline
			IBD	&armchair	&fireplace	&inside arm	&shade	&sofa	&frame	&back pillow\\ \hline
			\multicolumn{8}{c}{}\\
			
			\multicolumn{8}{l}{\textbf{Class: market/indoor}}\\ \hline
			Proposed	&pedestal	&sales booth	&shop	&case	&bag	&bulletin board	&food\\ \hline
			IBD	&sales booth	&pedestal	&food	&\cancel{fluorescent}	&shop	&shops	&apparel\\ \hline
			\multicolumn{8}{c}{}\\
			
			\multicolumn{8}{l}{\textbf{Class: soccer\_field}}\\ \hline
			Proposed	&grass	&pitch	&grandstand	&court	&person	&post	&goal\\ \hline
			IBD	&pitch	&field	&\cancel{cage}	&\cancel{ice rink}	&\cancel{tennis court}	&grass	&\cancel{telephone booth}\\ \hline
			\multicolumn{8}{c}{}\\
			
			\multicolumn{8}{l}{\textbf{Class: forest/broadleaf}}\\ \hline
			Proposed	&tree	&bush	&trunk	&\cancel{cactus}	&brush	&\cancel{fire}	&leaves\\ \hline
			IBD	&tree	&trunk	&bush	&leaves	&\cancel{semidesert}	&\cancel{grid}	&\cancel{clouds}\\ \hline
			\multicolumn{8}{c}{}\\
			
			\multicolumn{8}{l}{\textbf{Class: art\_school}}\\ \hline
			Proposed	&person	&hand	&paper	&\cancel{plaything}	&fabric	&bag	&board\\ \hline
			IBD	&paper	&drawing	&\cancel{plaything}	&painting	&hand	&board	&figurine\\ \hline
			\multicolumn{8}{c}{}\\
			
			\multicolumn{8}{l}{\textbf{Class: dining\_hall}}\\ \hline
			Proposed	&drinking glass	&stool	&table	&spindle	&menu	&person	&plate \\ \hline
			IBD	&plate	&light	&stool	&sash	&napkin	&display board	&spindle \\ \hline
			\multicolumn{8}{c}{}\\
			
			
			\multicolumn{8}{l}{\textbf{Class: butte}}\\ \hline
			Proposed	&mountain	&hill	&desert	&badlands	&rock	&valley	&land \\ \hline
			IBD	&hill	&badlands	&desert	&cliff	&\cancel{cloud}	&mountain	&\cancel{diffusor} \\ \hline
			\multicolumn{8}{c}{}\\
			
			\multicolumn{8}{l}{\textbf{Class: canyon}}\\ \hline
			Proposed	&mountain	&rock	&cliff	&hill	&badlands	&land	&desert \\ \hline
			IBD	&cliff	&mountain	&badlands	&desert	&pond	&\cancel{bumper}	&hill \\ \hline
			\multicolumn{8}{c}{}\\
			
			\multicolumn{8}{l}{\textbf{Class: coast}}\\ \hline
			Proposed	&sea	&sand	&land	&embankment	&rock	&\cancel{mountain}	&water \\ \hline
			IBD	&sea	&wave	&land	&\cancel{mountain pass}	&sand	&cliff	&\cancel{cloud} \\ \hline
			\multicolumn{8}{c}{}\\
			
		\end{tabular}
		\caption{Top seven concepts suggested by the proposed method and the IBD method.
			The strikethrough texts shows the errors where a concept that is irrelevant to the corresponding task class is selected.}
		\label{tab:explain}
	\end{table*}

\begin{figure}[!h]
		\centering
		\includegraphics[width=0.5\textwidth]{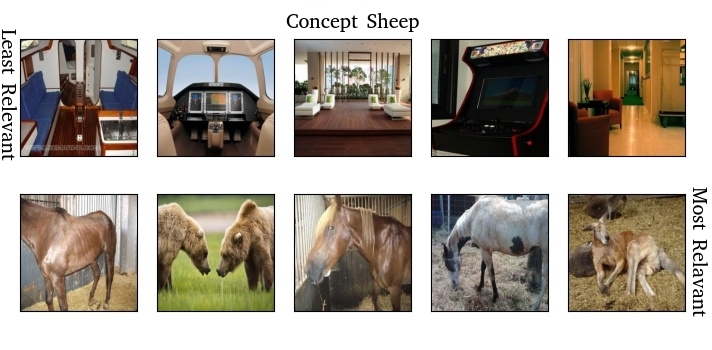}
		\includegraphics[width=0.5\textwidth]{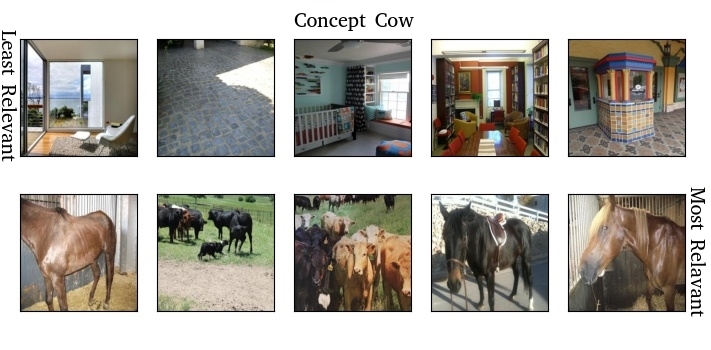}
		\includegraphics[width=0.5\textwidth]{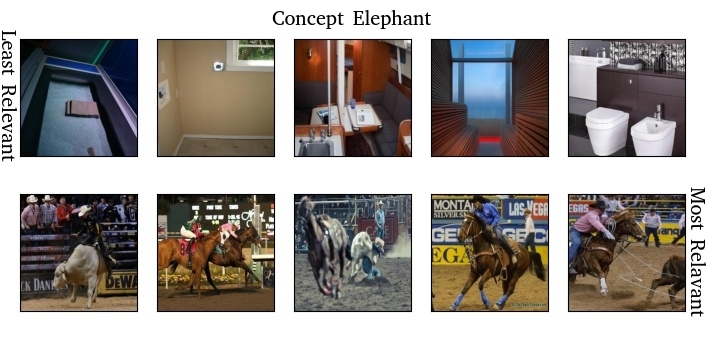}
		\caption{Some minimal and maximal samples of animal related concepts.
		These samples show that without a quality control step, the concepts can capture irrelevant information.}
		\label{fig:sortingProblem1}
	\end{figure}

	We also realized that the quality of the IBD benchmark concepts is not verified.
	So we sorted the samples in the distribution sample set according to the output of each concept classifier network.
	Figures \ref{fig:sortingProblem1} shows the results for concepts sheep, cow and elephant.
	For each concept, the first row shows the five images of the minimal concept classifier output in the increasing relevance order from left to right.
	In the second row, the five images that maximize the concept are shown in the increasing relevance order from left to right.
	This test acts as a sanity check on concept learning at a particular layer and shows how a concept may look like in that layer and may reveal if the concept is not reliable.
	For instance, we found that the concepts sheep, cow, and elephant are not reliable and maximized whenever a dirt field is present, (for example a horse race track) as shown in Figure \ref{fig:sortingProblem1}.
	The IBD method may blindly use such unreliable concepts in its explanations, but the proposed method verifies the accuracy of the concept classifiers and avoids using such concepts.
	A closer look at the second row, reveals that some images maximize more than one concept, which implies that the concept network cannot even distinguish between these concepts.
	In other words, in this layer, these individual animal concepts are merged and have formed a higher-level ``animal'' concept while the original concepts are forgotten (apparently the type of animal is not important for the Places365 classes).
	Usage of concepts without verification degrades the explanations of the IBD method.
	For instance, in IBD's explanation for the task class raft, the elephant concept appeared as the 5th most important concept.
	A similar phenomenon is observed for the bathroom-related concept bidet where it appeared in the IBD's fifth and sixth concepts of the staircase and beauty salon classes.

	\subsection{Concept Hierarchy Analysis}
	In this section, we demonstrate the proposed method for finding a hierarchy of concepts on two applications: scene classification and	prediction of Osteo-arthrosis severity from X-ray images.
	
	\subsubsection{Learning Concept Hierarchy for Scene Classification}
\begin{figure}
		\centering
		\includegraphics[width=0.35\textwidth]{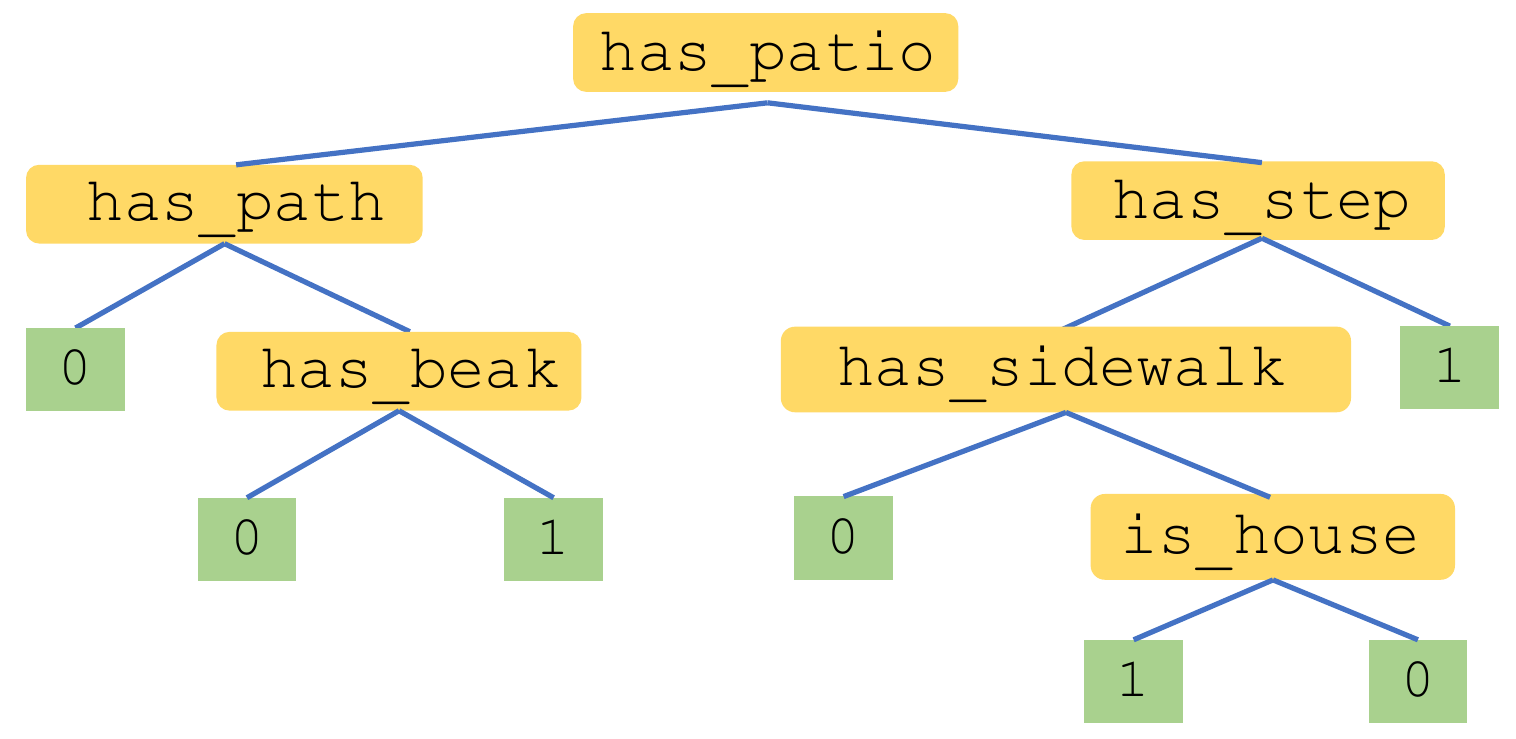}
		\caption{Concept  hierarchy  found  by  the  proposed  method  for  the  task class of zen garden.}
		\label{fig:DT-ZenGarden}
	\end{figure}

	In this experiment, we examine the neural network trained on places365 dataset \cite{zhou2017places} and try to explain task decisions based on the concepts provided in the Broden dataset \cite{Bau2017}.
	We test the proposed approach under two scenarios.
	In the first scenario, a decision tree is trained for explaining each task class.
	In the second scenario, a multi-class decision tree is trained for explaining the neural network as a whole.
	Figure \ref{fig:DT-ZenGarden} shows the decision tree for the ``zen garden'' class.
	Using this tree we can easily see what concept combinations correspond to this class and what concepts' presence can stop the classifier from labelling as ``zen garden''.
	We also trained a multi-class decision tree to predict the multi-class decisions of the entire classifier.
	Since the full description of the classifier is too long, a branch of the resulting decision tree is provided in the supplementary materials.

	\subsubsection{Learning Concept Hierarchy for Predicting Osteo-arthrosis from X-ray images}
\begin{figure}
		\centering
		\includegraphics[width=0.2\textwidth]{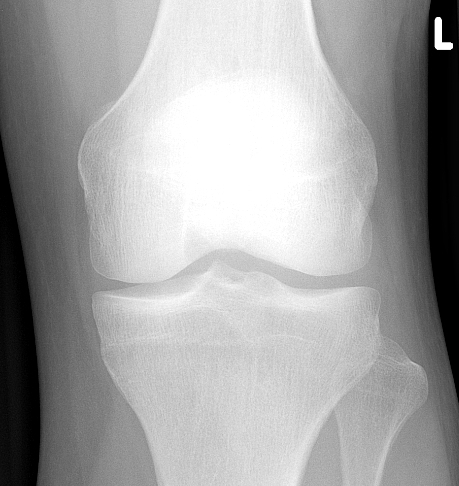}
		\caption{Sample of a knee X-Ray input image for KLG classifier.
		The letter L indicates that this is the image of the patient's Left leg.}
		\label{fig:xray}
	\end{figure}

	In this experiment, we examine a neural network trained to classify Osteo-arthrosis KLG score \cite{kellgren1958osteo}.	
	The KLG score is a 5-grade integer value between 0 and 4 that measures the severity of Osteoarthrosis in joints.
	We follow the preprocessing procedure mentioned in \cite{pierson2019using} and used in \cite{koh2020concept}.
	We use the OAI dataset \cite{nevitt2006osteoarthritis} to train a Resnet18 model to classify KLG scores based on the knee joint's X-ray images.
	OAI (Osteo-Arthrosis Initiative) is a data-gathering initiative from multiple sources on Osteo-Arthrosis disease.
	The collection includes X-Ray, MRI, etc images of different joints of patients taken every six months for up to 96 months.
	Labels for this dataset were created by the aggregating experts' opinions.
	Here we use only X-Ray knee images that have concept annotations and KLG score (16249 images).
	Following \cite{pierson2019using}, we use only concepts for which a minimum number of samples are presented in the dataset. 
	The X-ray images are reshaped into $512\times 512$ pixels.
	Then the task classifier is examined against concepts to see whether and how the task classifier is influenced by these concepts.
	In other words, whether the task classifier is following the expert's way of determining KLG scores.
	Note that all patient information has been removed by OAI and only the leg side (L/R) is remained on the image (see Figure \ref{fig:xray}).


\begin{figure}
		\centering
		\includegraphics[width=0.48\textwidth]{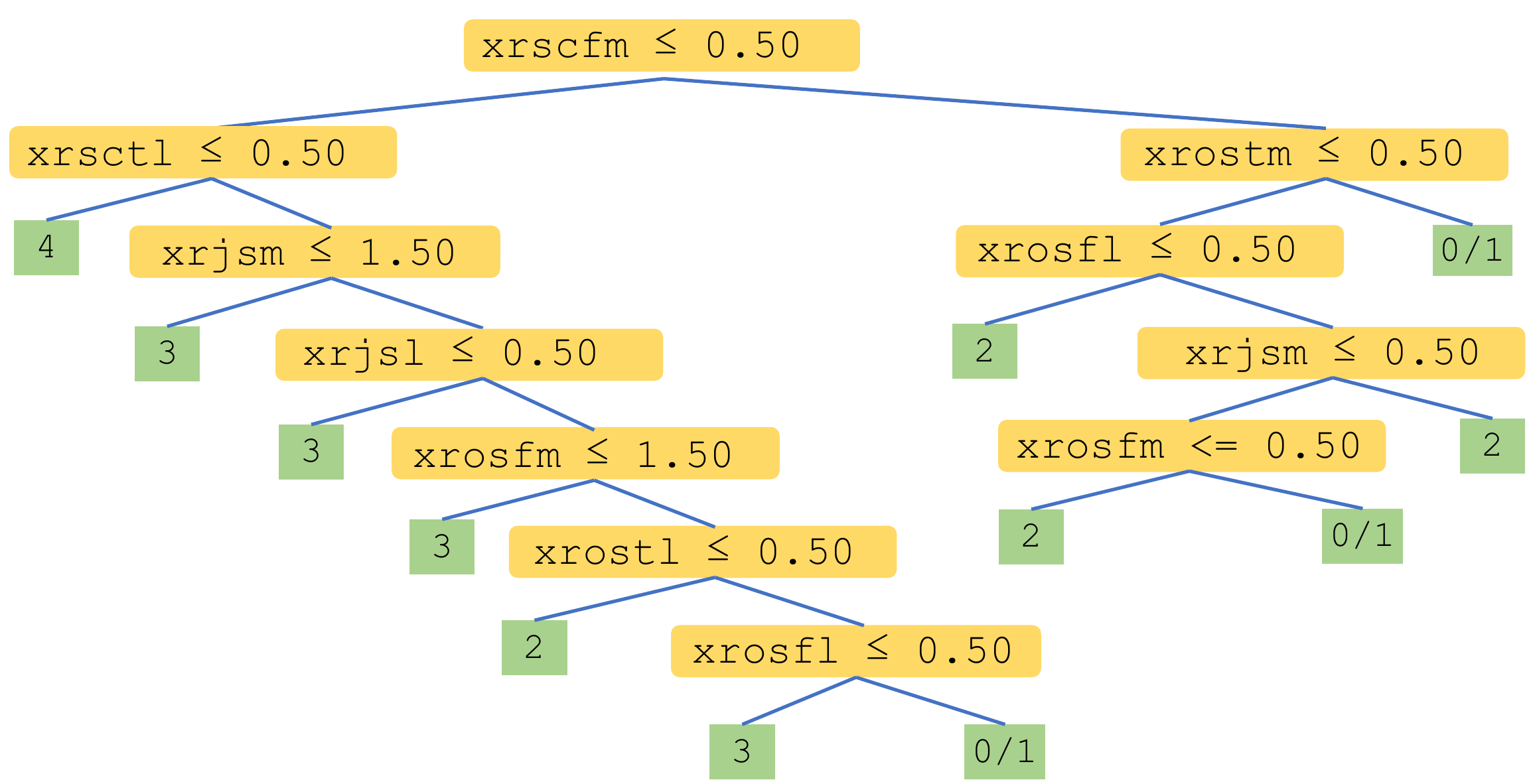}
		\caption{Concept hierarchy found by the proposed method for the neutral network that predicts Osteo-arthrosis KLG score.}
		\label{fig:DT-KLG}
	\end{figure}

	To prevent overfitting, we hard-decision the predicted concept values using some severity thresholds and consequently, concepts become binary values.
	Also, we stop splitting a node as soon as there are less than 2 samples per class in that node.
	We also merge back the leaves that have similar labels.
	The resulting decision tree is shown in Figure \ref{fig:DT-KLG}.
	In general, the presence of concepts usually is indicative of a more severe condition and hence a higher KLG score.
	The resulting decision tree is consistent with that rule-of-thumb in that at each node whenever we take a left branch -- corresponding to when the corresponding concept is higher than a threshold-- usually we end up in a leaf with a relatively higher KLG score.
	
	\section{Discussions}
	
	The process of learning a hierarchy of concepts involves two steps: first, learning a model for predicting each concept, and then representing the original network by a decision tree.
	How faithful is the decision tree to the model behaviour is determined by both the accuracy of the decision tree and concept networks.
	
	The distribution sample set represents likely cases of input and should be a good representation of the inputs that the network will be tested on.
	For instance, using the job title classification example, if we expect most but not all doctors to wear lab coats, it should be reflected in the distribution sample set.
	However, a systematic change, for example, if a large portion of the distribution sample set has a certain relationship, is expected to affect the resulting AUCs.
	For example, if we do not expect the network to be tested on images of physicians with a stethoscope in a garage fixing cars, the distribution sample set should not be dominated with such images.
	The fact that the distribution sample set does not need any kind of labelling, enables us to use any set of inputs like a held-out part of data or even inputs recorded from other sources, as long as they are a good representation of likely task classification inputs.
	Note that most of the methods that predict the behaviour of a network need a kind of sample set, for instance, TCAV \cite{Kim2018} needs samples from the task class.
	
	The choice of which layer to inspect is not a straightforward decision.
	The inspection of later layers is computationally cheaper (since the training of the concept classifier is cheaper).
	But there is no guarantee that the concepts are still present in those layers since the network might have traded them with a combination of concepts more useful for the task classification.
	For instance presence of a stethoscope might not be detectable in the last layer, but the presence of a medical instrument might be possible (distinguishing images that include a medical instrument from the ones that don't).
	For this reason, we start our analysis from the last layer in the network and work our way back until we reach a layer that the concept is present (good accuracy of concept network) or reach the first layer which would indicate that the concept is too complex to be detected by the network.

	\section{Conclusion}
	We proposed a framework for verifying the presence of human-friendly concepts in activations of intermediate layers of a neural network.
	The proposed method quantitatively determines the causal relationship between a concept and the neural network task classes.
	We also determined the interactions between concepts with respect to a task class by constructing a hierarchy of concepts in the form of a decision tree.	
	We showed the effectiveness of the proposed methods through several comparative experiments on synthetic and real-world datasets, demonstrating improved performance compared with the previous methods.
	
	
	


	\section*{Acknowledgement}
	This work was supported by Natural Sciences and Engineering Research Council, Vector Institute and Compute Canada.

	\bibliography{ref}
	\bibliographystyle{IEEEtran}

\end{document}